\documentclass[conference]{IEEEtran}
\IEEEoverridecommandlockouts
\usepackage{cite}
\usepackage{verbatim}
\usepackage{amsmath,amssymb,amsfonts}
\usepackage{algorithmic}
\usepackage{graphicx}
\usepackage{textcomp}
\usepackage{flushend}
\usepackage{algorithm}
\usepackage{multirow}
\usepackage{array}
\usepackage[lofdepth,lotdepth]{subfig}

\usepackage{graphicx}
\usepackage{placeins}

\usepackage[colorinlistoftodos]{todonotes}
\usepackage[colorlinks=true, allcolors=blue]{hyperref}

\usepackage{booktabs}
\usepackage{multirow}
\usepackage{textcomp}

\def\BibTeX{{\rm B\kern-.05em{\sc i\kern-.025em b}\kern-.08em
    T\kern-.1667em\lower.7ex\hbox{E}\kern-.125emX}}
\usepackage{authblk}
\title{Compressively Sensed Image Recognition}
\author[1]{Ay\c{s}en De\v{g}erli}
\author[2,3]{Sinem Aslan}
\author[1]{Mehmet Yama\c{c}}
\author[4]{B\"{u}lent Sankur}
\author[1]{Moncef Gabbouj}
\affil[1]{Tampere University of Technology, Laboratory of Signal Processing, Tampere, Finland}
\affil[2]{Ca\textquotesingle \hspace{0.1cm}Foscari University of Venice, European Centre for Living Technology, Venice, Italy}
\affil[3]{Ege University, International Computer Institute, \.{I}zmir, Turkey}
\affil[4]{Bo\v{g}azi\c{c}i University, Electrical and Electronics Engineering Department, \.{I}stanbul, Turkey}
\begin{document}

\maketitle

\begin{abstract}
Compressive Sensing (CS) theory asserts that sparse signal reconstruction is possible from a small number of linear measurements. Although CS enables low-cost linear sampling, it requires non-linear and costly reconstruction. Recent literature works show that compressive image classification is possible in CS domain without reconstruction of the signal. In this work, we introduce a DCT base method that extracts binary discriminative features directly from CS measurements. These CS measurements can be obtained by using (i) a random or a pseudo-random measurement matrix, or (ii) a measurement matrix whose elements are learned from the training data to optimize the given classification task. We further introduce feature fusion by concatenating  Bag of Words (BoW) representation of our binary features with one of the two state-of-the-art CNN-based feature vectors. We show that our fused feature outperforms the state-of-the-art in both cases.
\end{abstract}


\section{Introduction}

The first step in any signal processing task is the acquisition of signals. The classical pathway for band-limited signals is to instantaneously sample the signal at the Nyquist-Shannon rate, then compress the signal to remove redundancies and/or irrelevancies, typically using a transform-based compression technique for efficient storage and transmission. The compressed signal must be decompressed before executing any further signal processing operation such as classification, detection, inference. etc.  %. if the signal is sparse or compressible in a proper domain; (iii) at the receiver side, the compressed signal is decompressed before any signal processing operation for a computer vision task, e.g. classification, detection etc. Since the amount and dimension of the digital samples are getting increased every day as a result of technological developments, it has been almost necessity to apply dimensionality reduction, e.g. by employing PCA \cite{Classical PCA}, random projection \cite{RandomProjection}, etc., before making any inference on the decompressed data \cite{DataDeluge}.  %. Because the amount and dimension of the digital samples are getting increased everyday as a result of technological developments \cite{DataDeluge}. As a result, it has been important for the signal processing community to develop dimensionality reduction techniques such as PCA \cite{Classical PCA}, random projection \cite{RandomProjection}, etc., for many machine learning and signal processing tasks.  

The new sampling paradigm, \textit{Compressive Sensing} (\textit{CS}) \cite{CS} bypasses this laborious Nyquist-Shannon data acquisition scheme in that signals are being compressed while being sampled with random patterns. Thus the sampling and compression steps are combined into one action. However, the  reconstruction of the signal from \textit{compressively sensed measurements} (\textit{CSMs}) becomes  non-linear and considerably costlier in the computational effort. This costly signal reconstruction operation would be counterproductive  %as mentioned above in many signal analysis and machine learning applications dimension reduction is
were it not for the emerging signal processing algorithms in the compressed domain. A newly emerging idea \cite{CS-Classification1,CS-classification2,signalPro} is using CSMs directly in inference problems without executing any reconstruction. This promising approach can potentially be advantageous in real-time applications and/or when dealing with big data. 

%Although the inference problems such as detection, recognition etc., are well studied in the literature for the signals acquired by traditional Nyquist based methods, the inference problems on signals in CS domain has been recently emerging active research area. 
In a pioneering work, Davenport et al. \cite{signalPro} have addressed the problem of inference directly on compressively sensed measurements. In \cite{CS-Learning}, it is theoretically shown that the accuracy of the soft margin SVM classifier is preserved when data is collected with sparse random projections. %random projection is done and data has a sparse representation. They also attempt to give some tight bounds. 
The authors in \cite{smashedfilter1} have introduced the idea of smashed filter and, based on the Johnson-Lindenstrauss Lemma \cite{Johnson} have shown that the inner product of two signals is relatively preserved for compressively sampled signals when the sampling matrix consists of random values, chosen from some specific probability distributions. Different versions of the smashed filter are used in various applications \cite{smashedfilter2,smashedfilter3}. For instance in \cite{smashedfilter3},  a compressive smashed filter technique is proposed by first producing a set of correlation filters from uncompressed images in the training set, and then at the testing stage by correlating CSMs of test images and with the CSMs of learned filters. A linear feature extraction method in CS domain is developed in \cite{EEG1} for direct classification of compressively sensed EEG data. In  \cite{ECG} fed an SVM classifier with a fusion of CSMs (projected data) and dynamic features and they reported performance beyond the state-of-the-art for $1$-D ECG data classification. 

Other works, e.g., \cite{bounds1,bounds2}, have provided theoretical guarantees for achievable accuracy in different CSM setups for both sparse and non-sparse cases.

%\textbf{[CNN'in kullanildigi calismalarin onemi? ne gibi basarimla elde etmisler bu problem icin? ]}

All the above works try to use compressed samples directly to solve the inference problem. A second approach is to boost the size of CSMs to the original image size by a simple linear projection, but avoiding the costly nonlinear reconstruction procedure. This simple back-projection yields a pseudo-image and one then proceeds with the inference task on this imperfectly reconstructed image. This image, restituted to its original dimension and also known as \emph{proxy image}, is usually a heavily degraded version of the original image. One way to obtain the proxy image is by premultiplying the compressed image by the transpose of the sampling matrix. In \cite{cnn}, the authors apply a CNN-based feature extraction method on such a  proxy image. Their measurement matrix consists of random Gaussian distributed numbers.   Another work \cite{deeplearning} uses a deeper network structure (as compared to \cite{cnn}) by adding two fully connected layers at the beginning of the network. Thus this network can learn as well the linear dimension reduction (so-called measurement matrix) and linear back projection to the image domain (i.e., the transpose of the measurement matrix).

In this work, following the vein of the second approach we propose a DCT-based discriminative feature scheme, computed directly from the proxy image. This feature vector (called MB-DCT) is binary, hence simple and low cost. A preliminary version of this feature was presented in EUSIPCO \cite{sinem}. In this work, we applied MB-DCT on non-compressively sampled images. In  \cite{sinem}  we had shown that this simple scheme of selected binarized DCT coefficients, computed in increasing scales of local windows was remarkably robust against linear and nonlinear image degradations, such as additive white Gaussian noise, contrast and brightness changes, blurring, and strong JPEG compression. We %modify
use MB-DCT scheme in \cite{sinem} for feature extraction from image proxies. Our experimental results show that using this simple binary feature method surpasses the performance of Smashed Filters \cite{smashedfilter3}. We further introduce feature fusion by concatenating Bag of Words (BoW) representation of our binary features with one of the two state-of-the-art CNN-based feature vectors, i.e., in \cite{cnn} and \cite{deeplearning}. In the method \cite{cnn}, elements of measurement matrices were drawn from a random distribution as typical in conventional CS theory, whereas in the method \cite{deeplearning} sampling matrices were learned from a deep network; the latter method proved to be superior for smaller measurement rates. However, random sensing scheme may still be needed for some applications where one needs to pre-classify the data directly using CSMs, then reconstruct the signal for further analysis. For instance in a remote health monitoring system, we may wish to detect anomalies directly from CSMs of ECG signal on the sensor side. Then based on the sensor side classification, CSMs of selected cases can be transmitted for a more detailed analysis by a medical doctor. Therefore, we consider the random sensing approach and learned sensing approach as two different set-ups. In this paper, we show that our fused features outperforms the aforementioned works for both of the schemes and gives the state-of-the-art performance.

%In this work, we propose to use a modified version of an image descriptor, namely MB-DCT, for direct inference of compressive sensed images. Examined for image recognition problem, it is demonstrated in \cite{sinem} that MB-DCT is highly robust against linear and nonlinear image degradations such as Additive White Gausian Noise, contrast and brightness changes, blurring, and spatial quantization as a result of JPEG compression. Convinced with the highly successful results reported in \cite{sinem}, we employed  MB-DCT for classification of CS proxy that are pseudoimages incurred by linear degradation (see Eq. \ref{Eq4}). Specifically, we use MB-DCT in two main schemes:  1) in the conventional BoVW (Bag of Visual Words) framework as employed in \cite{sinem}; 2) in the fusion scheme for complementing deep learning features. We present the details of the work in the following subsections.

We briefly introduce the notation used and some preliminary information. We define the $\ell_0$-norm of the vector $x \in \mathbb{R}^N$ as $\left \| x \right \|_{\ell_0^N} = \lim_{p \to 0} \sum_{i=1}^N \left \vert x_i \right \vert^p = \# \{ j: x_j \neq 0 \}$. The compressive sensing (CS) scheme extracts $m$ number of measurements from the N-dimensional input signal $S \in \mathbb{R}^N$, i.e.,
\begin{equation}
 y= \Psi S,
\end{equation}
where $\Psi$ is the $m \times N$ measurement matrix and typically $m << N$. Consider this signal to be $k$-sparse in a sparsifying basis $\Phi$ such that $S=\Phi x$ with $\left \| x \right \|_{\ell_0^N} \leq k$. Then, the general compressive sensing setup is
\begin{equation}
y = \Psi \Phi x = A x,
\end{equation}
where $A = \Psi \Phi$ is sometimes called as holographic matrix.  It has been demonstrated that the sparse representation
in \eqref{sparse_rep} is unique if $m \geq 2k$ \cite{nullspace}.
\begin{equation}
\min_x ~ \left \| x \right \|_{\ell_0^N}~ \text{subject to}~ Ax=y \label{sparse_rep}
\end{equation}

The organization of the rest of the paper is as follows. In Section II, we provide the notation, mathematical foundations and a brief review of CS theory. The difference between the two measurement approaches, namely, based random weights or learned weights in the acquisition of CMSs and reconstruction of proxies are explained in Section III. Then in Section IV, we introduce the proposed feature extraction method from the two proxy varieties. Finally, performance evaluations of the proposed method are given and a conclusion is drawn.

%\section{Preliminaries} We define the $\ell_0$-norm of the vector $x \in \mathbb{R}^N$ as $\left \| x \right \|_{\ell_0^N} = \lim_{p \to 0} \sum_{i=1}^N \left \vert x_i \right \vert^p = \# \{ j: x_j \neq 0 \}$. 

%\subsection{Compressive Sensing}
%Compressive Sensing (CS) asserts that one can recover a sparse signal from far fewer measurements than the traditional methods use. Mathematically speaking, we have $m$ number of measurements of $S \in \mathbb{R}^N$, i.e.,
%\begin{equation}
% y= \Psi S,
%\end{equation}
%where $\Psi$ is the $m \times N$ measurement matrix. Consider this signal is $k$-sparse in a sparsifying basis $\Phi$ such that $S=\Phi x$ with $\left \| x \right \|_{\ell_0^N} \leq k$. Then, the general compressive sensing setup is
%\begin{equation}
%y = \Psi \Phi x = A x,
%\end{equation}
%where $A = \Psi \Phi$ is sometimes called as holographic matrix.  Using the elementary linear algebra one can say that the sparse representation
%in \eqref{sparse_rep} is unique if $m \geq 2k$ \cite{nullspace}.
%\begin{equation}
%\min_x ~ \left \| x \right \|_{\ell_0^N}~ \text{subject to}~ Ax=y \label{sparse_rep}
%\end{equation}

\section{Related Works}

The signal reconstruction expounded in \eqref{sparse_rep} is an NP-hard problem. Among the plethora of methods to overcome the computational impasse one can list convex relaxation, various greedy algorithms, Bayesian framework, non-convex optimization, iterative thresholding methods etc. \cite{justrelax}. However, these algorithms still suffer from computational complexity and must be restricted mostly to non-real time applications. For an application where a fast and real-time data inference is required, one possible solution could be designing a non-iterative solution such as a simple forward pass re-constructor \cite{Learningtoinvert,Learningtoinvert2} based on neural networks. These types of solutions, nevertheless, remain still wasteful of resources since we have to return to the high-dimensional ambient domain from the compressed domain in order to execute tasks such as feature extraction, classification etc. Furthermore the exact recovery probability, that is the phase diagram of the recovery algorithms, depends critically on the sparsity level $k$ and the number of measurements, $m$ \cite{DAMP}.  When the  proportion of measurements is very low, typically for $\frac{m}{N} \leq 0.1 $ most reconstruction algorithms fail. Approaches to tackle the reconstruction bottleneck have been to bypass the reconstruction step altogether, and make inferences directly on the sparse signal $y$ \cite{smashedfilter3}, or on some  proxy of the signal, $\tilde{S}=\Psi^T y$ without solving the inverse problem for  sparse reconstruction $\hat{x}$ as in Eq. \eqref{sparse_rep}, therefore $\hat{S}=\Phi \hat{x}$, where $\hat{S}$ full recovery of the vectorized image. We can express the linear degradation on the proxy as 
\begin{equation}
 \tilde{S} = \Psi^T y = \Psi^T \Psi S = H S,  
\label{Eq4}
\end{equation}
where $H=\Psi^T \Psi$ is a non-invertible matrix that represents the non-linear degradation on original signal $S$.
\subsection{Feature extraction from compressively sensed signals with random measurement matrices}
\begin{figure}[H]
\centering
\includegraphics[width = \linewidth]{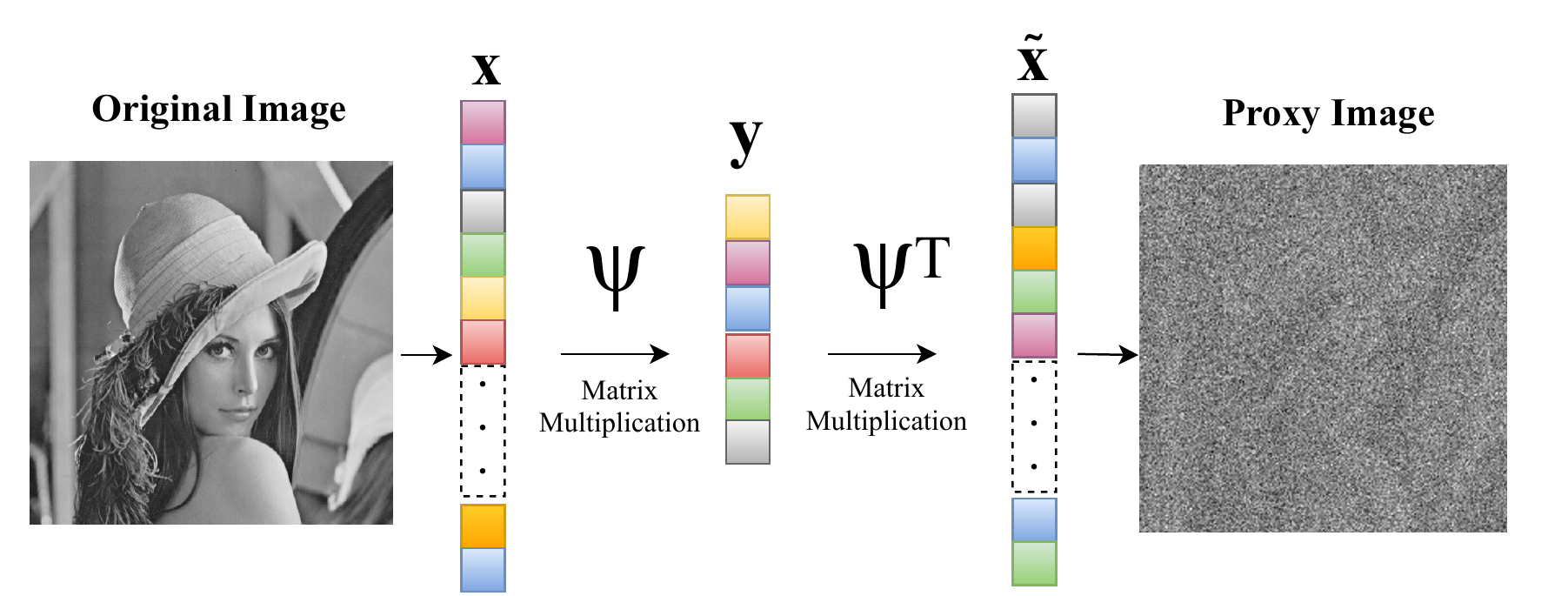}
\caption{The image of Lena and its proxy $\Psi^T y$ obtained from CSMs, where $y$ results from projecting the original image on a Gaussian random measurement matrix.}
\label{fig:fgr1}
\end{figure}

In order to guarantee the exact recovery of the $k$-sparse signal $x$ from $y$, the measurement matrix $\Psi$ should satisfy certain properties. For example, the measurement matrix, $\Psi$ with i.i.d. elements $\Psi_{i,j}$ drawn according to $\mathcal{N}\left ( 0,\frac{1}{m} \right )$, and $m > k (\log (N/k))$ guarantees with high probability the exact signal reconstruction when we relax the $\ell_0$ to $\ell_1$ in \eqref{sparse_rep} \cite{ca08}. Random measurement matrices are known to be universally optimum in the sense that they are data independent of characteristics of the data, and they satisfy minimum reconstruction error with minimum $m$ when we do not have another prior information about $k$-sparse signal. The acquisition of the proxy signal is obviously done as, $\tilde{S} = \Psi^T y$, where $\Psi^T \in \mathbb{R}^{N \times m }$ is the transpose of the measurement matrix $\Psi$. An example proxy image is shown in Figure \ref{fig:fgr1}. 
\subsection{Feature extraction from compressively sensed signals based on measurement matrices with learned coefficients}

\begin{figure}[H]
\centering
\includegraphics[width = \linewidth]{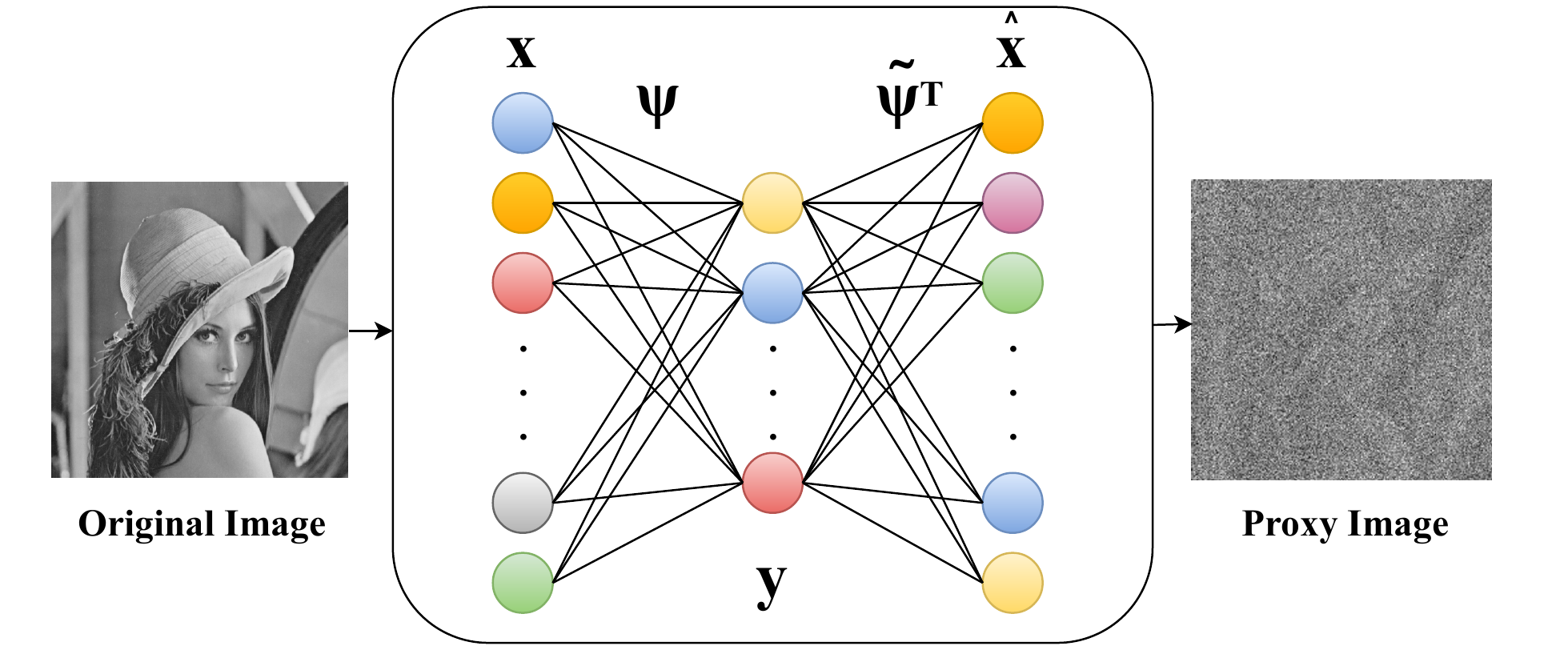}
\caption{An  original image and its proxy $\Psi^T y$ where $y$ that is obtained using the learned measurement matrix.}
\label{fig:fgr3}
\end{figure}

Design of optimal measurement matrices for CS reconstruction and/or for inference tasks is an active research area. An approach to learn a  projection operator from image to measurement domain and its backprojection operator from compressed domain to image domain is presented in \cite{cnn}. The authors have used two fully-connected layers that are followed by convolution layers. The first layer takes the original image $S$ and projects it to the measurement domain, $y$. The learned weights of this layer represent the elements of measurement matrix for compressively sensing images. The second layer represents the back projection to the image domain to produce a proxy of the image, i.e., $\tilde{S} = \tilde{\Psi^T} y$. In this expression $\tilde{\Psi^T}$ is the learned transpose of the measurement matrix, which is used instead of the transpose of the true measurement matrix, $\Psi^T$. The output of this layer, the proxy image, is given as input to convolutional layers to realize some non-linear inference task, e.g., classification. Thus the measurement matrix, the pseudo-transpose of the measurement matrix and convolutional network are all jointly learned from the training data. Figure \ref{fig:fgr3} illustrates the first two fully-connected layers of this network.
\vspace{-0.5cm}

\section{Proposed Approach}
In this work, we have employed %modified 
MB-DCT features extracted from proxy images (see Eq. \ref{Eq4}) for classification tasks. Specifically, we applied the MB-DCT of \cite{sinem} as follows: i) we use 4 window scales (instead of 6 as in \cite{sinem}) and the size of the largest window is now 24 pixels instead of 128 pixels \cite{sinem}) to fit smaller sized MNIST images ($28 \times 28$ pixels); ii) we apply a different scheme of coefficient elimination in that we keep the best performing of the three sets (based on AC energy preservation) of DCT coefficients in the sense of classification accuracy.  An MB-DCT descriptor consists of mean quantization of 2D-DCT transform coefficients as computed from multiscale blocks around (densely or sparsely chosen) image points \cite{sinem}. 

We employ the mentioned scheme of MB-DCT features in two main modes:  1) The conventional BoW framework as  in \cite{sinem}; 2) A fusion scheme where MB-DCT features are complemented with deep learning features. 

\subsection{MB-DCT}
We review briefly the MB-DCT features:

\textit{(1) DCT computation:} 2D-DCT coefficients are computed in multiple nested blocks around selected image points, each incrementally changing in size. Similar to \cite{sinem}, we employ various sized windows in this work, in order to capture contextual information in different sized neighbourhoods around every image point. We compute 2D-DCT in four scales corresponding to block sizes \{8, 12, 16, 24\}, which seemed adequate for $28\times28$ pixel-sized MNIST images. For larger images, larger block sizes can be investigated for performance-computational cost tradeoff.

\begin{figure*}[h!]
\centering
\includegraphics[width = 1\textwidth]{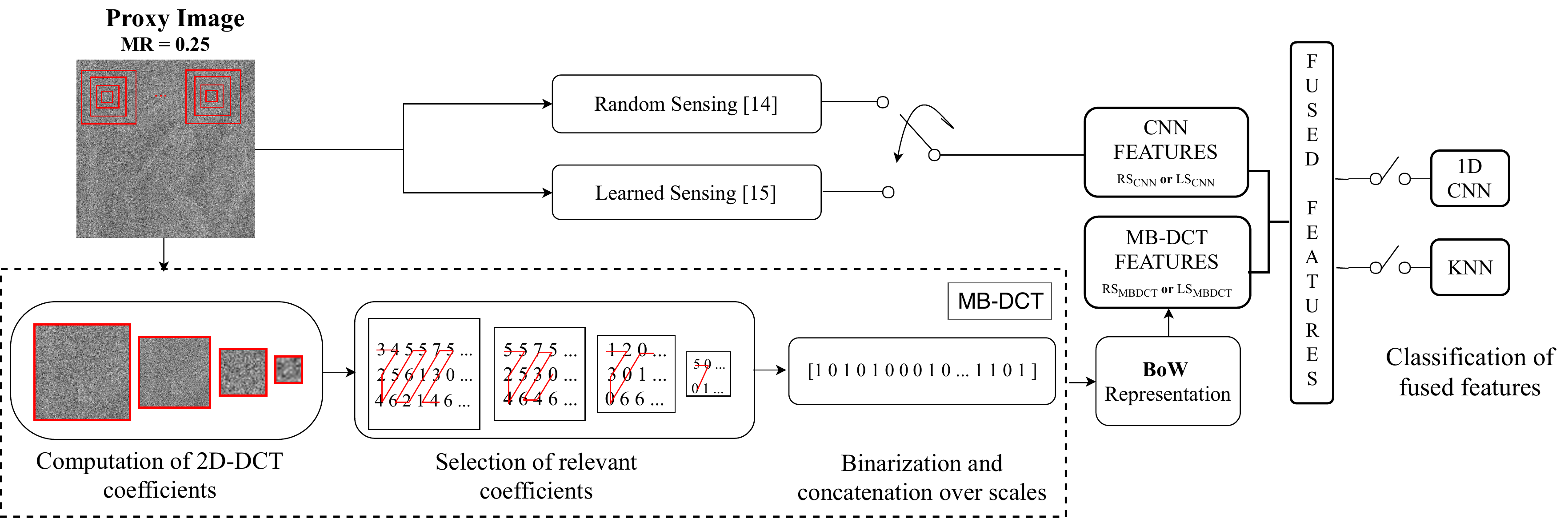}
\caption{Computational pipeline for the proposed approach.}
\label{fig:fgr4}
\end{figure*}

\textit{(2) Eliminating irrelevant coefficients:} A subset of zig-zag ordered DCT coefficients are kept for each block as features and the remaining coefficients are eliminated as irrelevant. The DC term was discarded in all scales as in \cite{sinem}, which also desensitizes the feature vectors to illumination level. We experimented for different sized subsets of the zig-zag ordered coefficients in each scale.  Specifically, to determine the quantity of DCT coefficients kept, we start with a random subset of training images. Then, we find three sets of zig-zag ordered DCT coefficients for every pixel location preserving, respectively, 90\% and 95\% of the AC energy. We repeat this experiment for each scale and for each window size. Finally, we fix the number of coefficients for each energy level (and for each scale)  to the average, over all training images,  number of coefficients that have met energy preservation percentages. For 100\% of the energy preservation we keep all the AC coefficients. For the window sizes of $\{8,12,16,24\}$, we found that the average number of AC coefficients corresponding to 90\% and 95\% %and 100\% 
energy are $\{15, 26, 37, 73\}$ and $\{21, 40, 63, 130\}$, %\textcolor{red}{\%100 icin sayi vermeye gerek var mi? Zaten boyun karesi - 1 olacaktir, yani tum AC katsaylari}$\{63, 143, 255, 575\}$
respectively. Using these sets of coefficients specific to the window size, we measured the classification error rate on MNIST proxy images at different sensing rates. The resulting error rates are given in Table 1, where one can see that  the performances do not differ significantly. Notice that one needs to use quite larger set of  coefficients as compared to  when original images were used  \cite{sinem}. Due to the imperfect reconstruction of proxy images, the structural information in the image is not as compact as in the original one. 
 
%While low-frequency DCT coefficients endure the predominant portion of the energy and have the sufficient discriminative information, they are also less sensitive to minor spatial perturbations \cite{bsankur}. Primarily, The DC term was discarded in all scales as in \cite{sinem}, which also desensitizes the feature vectors to illumination level. It is reasonable to discard the DC term because it does not posses much discriminative information \cite{bsankur} and desensitizes the feature vectors to illumination level. Discarding the DC term, 

\textit{(3) Binarization of the coefficients:} Since binary features are memory and computation efficient, we binarized the selected coefficients of each block by mean quantization similar to \cite{sinem}. We also tried median quantization and also trimming, but the mean quantization provide slightly better results (around 0.5\% improvement). 

\textit{(4) Concatenation of different scales:} The final binary descriptor for a given keypoint is obtained by concatenation of binarized DCT coefficient sets at each scale. 

The computational pipeline of the MB-DCT scheme is illustrated in Figure \ref{fig:fgr4}.
%textbf{[sa: sonunda kac bit lik feature'umuz oldu? -> 1000kusur bit uzunlugunda, onceki bildiride 192 ve 256 bit de kalmak icin  buyuk scale'ler secmis ama her scale de cok katsayi elimination i yapmistik. )}

%\textbf{sa: eski calismadaki feature ile bu verikumesinde sonuc alip, yeni mbdct nin faydasini gostermeliydik. camera ready'de artik.}

%\textbf{sa: ayrica MNIST e ek  olarak, yeni bir verikumesinde de performans gosterebiliriz.}

\begin{table}[b!]
\centering
\caption{Effect of the quantity of DCT coefficients, as a function of energy preserved,  on classification performance of MNIST proxy images at different measurement rates. %obtained on the proxy of compressive sensed MNIST images obtained in a variety of measurement rates. 
}
\label{tab1}
\begin{tabular}{cccc}
\hline
\hline
Measurement Rate & 90\% Energy & 95\% Energy & 100\% Energy \\ \hline
0.25 & 8.75 & 8.67 & 7.26 \\
0.10 & 10.81 & 10.57 & 9.49 \\
0.05 & 16.04 & 15.21 & 14.28 \\
0.01 & 41.99 & 41.1 & 41.33 \\ \hline
\end{tabular}
\end{table}

\subsection{Performance of MB-DCT features for classification of CS proxies}
We first compute  MB-DCT features densely on the CS proxies of input images as in Eq. \ref{Eq4}. Then, we extract image descriptors from these features according to the two schemes explained in the sequel. 
\subsubsection{MB-DCT in the BoW framework}
In this scheme, we follow the conventional BoW procedure to compute descriptors of CS proxies. We learn a visual dictionary by K-Means clustering of dense MB-DCT features using hamming distance and computed on a  training image set. The MB-DCT feature of each image point is assigned the nearest binary descriptor from the dictionary with hard voting. Finally, we apply average pooling to compute a single image signature to obtain the BoW representation of each image. 

\subsubsection{Fusion of MB-DCT with Deep Learning features} 
Deep learning approaches have been shown to provide superior performance in the solution of inference problems provided sufficient amount of training data is available. Nevertheless, recent studies have demonstrated that the joint use of learned features and hand-crafted features (e.g., MB-DCT) can result in improved performance \cite{danaci,aslan2017}. 

For this purpose, we have jointly used the BoW descriptors obtained from MB-DCT features with CNN features computed as in the two recent works, i.e. \cite{cnn} and \cite{deeplearning}.  In both cases, proxy images are recovered by pre-multiplying the CSM vector with the transpose of the sensing matrix. In  \cite{cnn}, the sensing matrix consists of random Gaussian numbers while in \textit{Compressive Learning} (\textit{CL})  the sensing matrix is obtained using a deep learning architecture. In both approaches, CNN features are computed on the proxy images. This procedure of MB-DCT and CNN features is shown in the two upper branches of the block diagram in Figure \ref{fig:fgr4}.   We have named the CNN-derived feature scheme in \cite{cnn} as \textit{Random Sensing + CNN} (shortly \textit{$RS_{CNN}$}) and that in \cite{deeplearning}  as \textit{Learned Sensing + CNN} (shortly \textit{$LS_{CNN}$}), respectively. 

\begin{figure}[h!]
\centering
\includegraphics[width = \linewidth]{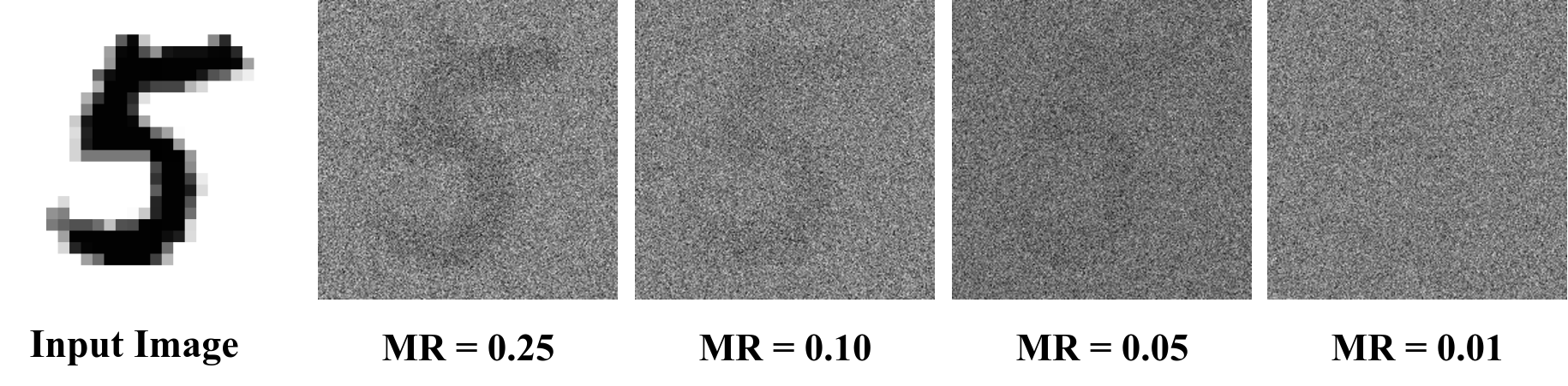}
\caption{Proxy images recovered when random sensing is used at different sensing rate}
\label{fig:fig5}
\end{figure}

Some examples of proxy images recovered with the transpose of the random Gaussian matrix using Eq. \ref{Eq4}  are shown in Figure \ref{fig:fig5} for four sampling rates. Starting from such a proxy image, we compute CNN features (coefficients of the fully connected last layer) using the Lenet5 model \cite{cnn}. We also compute in parallel  BoW descriptors from MB-DCT features, and we refer to this method as \textit{$RS_{MB-DCT}$}. %\textcolor{red}{Artilar, eksiler kafa karistiriyor. Onerim hepsini altcizgi karakteri yapmak, yani $RS_CNN, RS_MB_DCT ya da RS_MBDCT, LS_CNN$} \textcolor{brown}{ok.} . 
Finally, after  $L_2$ normalization, separately of each descriptor, we concatenate them to obtain the joint descriptor. We denote the fused descriptor as \textit{$RS_{(CNN|MB-DCT)}$}.

\begin{table*}[t!]
\centering
\caption{Test error rates on MNIST dataset. MR: Measurement Rate; RS: Random Sensing; LS: Learned Sensing, [$^\dagger$] denotes our re-implementation of \cite{cnn} and \cite{deeplearning}; [$^\ast$] denotes our proposed features. Presented results are obtained with the KNN classifier.} 
\label{tab2}
\begin{tabular}{c|c|cccc|cccc}
\hline
MR & 
\begin{tabular}[c]{@{}c@{}}Smashed\\Filter\cite{smashedfilter3}\end{tabular} & 
\begin{tabular}[c]{@{}c@{}}\scriptsize$RS_{CNN}$\cite{cnn}\end{tabular} & 
\begin{tabular}[c]{@{}c@{}}\scriptsize$RS_{CNN}^\dagger$ \end{tabular} & 
\begin{tabular}[c]{@{}c@{}}\scriptsize$RS_{MBDCT}^\ast$\end{tabular} &
\begin{tabular}[c]{@{}c@{}}\scriptsize$RS_{(CNN|MBDCT)}^\ast$\end{tabular} & 
\begin{tabular}[c]{@{}c@{}}\scriptsize$LS_{CNN}$\cite{deeplearning}\end{tabular} & 
\begin{tabular}[c]{@{}c@{}}\scriptsize$LS_{CNN}^\dagger$\end{tabular} & 
\begin{tabular}[c]{@{}c@{}}\scriptsize$LS_{MBDCT}^\ast$\end{tabular}& 
\begin{tabular}[c]{@{}c@{}}{}\scriptsize$LS_{(CNN|MBDCT)}^\ast$\end{tabular} \\ 
\hline

0.25 & 27.42\% & \textbf{1.63\%} & 1.73\%  & {7.26\%} & 2.17\% & \textbf{1.56}\% & 1.95\%& 5.84\% & {1.58}\% \\
0.10 & 43.55\% & \textbf{2.99\%} & 2.98\%  & {9.46\%} & 3.02\%  & 1.91\% & 1.88\%& 5.90\% & \textbf{1.58}\% \\
0.05 & 53.21\% & 5.18\% & 4.78\% & {14.28\%} & \textbf{4.44\%} & 2.86\% & 2.12\% & 5.80\% & \textbf{1.59}\% \\
0.01 & 63.03\% & 41.06\% & 45.8\%  & {41.33\%} & \textbf{24.78\%} & 6.46\% & 5.52\% & 19.88\%  & \textbf{3.87}\% \\ \hline
\end{tabular}
\end{table*}

\begin{table*}[t!]
\centering
\caption{Test error rates of the proposed features on MNIST dataset obtained with different classifiers}
\label{tab3}
\begin{tabular}{c|cccc|cccc}
\hline
\multirow{2}{*}{Measurement Rate} & 
\multicolumn{2}{c}{$RS_{MBDCT}$} & 
\multicolumn{2}{c|}{$RS_{(CNN|MBDCT)}$} & 
\multicolumn{2}{c}{$LS_{MBDCT}$} & 
\multicolumn{2}{c}{$LS_{(CNN|MBDCT)}$} \\ \cline{2-9} 
 & KNN & 1D-CNN & KNN & 1D-CNN & KNN & 1D-CNN & KNN & 1D-CNN \\ \hline
0.25 & 7.26\%  & 8.37\% & 2.17\%  &  1.69\%   & 5.84\% &5.88\% & 1.58\% & 1.58\%\\
0.10 & 9.46\% & 10.01\%  & 3.02\%   & 2.87\%  & 5.90\% & 6.19\% & \textbf{1.58\%} & {1.75\%}\\
0.05 & 14.28\%  & 14.16\%& \textbf{4.44\%} & 4.66\% & 5.80\% & 5.64\% & \textbf{1.59\%} & {1.63\%}\\
0.1 & 41.33\%  & 48.42\%&  \textbf{24.78\%} & 28.11\% & 19.88\% & 21.09\% & \textbf{3.87\%} & {4.57\%}\\
\end{tabular}
\end{table*}

For the $LS_{CNN}$ algorithm \cite{deeplearning}, we learned the sampling matrix for the MNIST dataset, i.e. we get the $\Psi$ and $\Psi^T$ matrices in Eq. \ref{Eq4} from the first and second fully connected layers of the trained network. We compute BoW representation of MB-DCT features on these proxies referred to as \textit{$LS_{MB-DCT}$}. Similarly, we get the CNN features from the last fully connected layer of the network. Finally, applying $L_2$ normalization to each, we concatenate them to obtain the joint features that we name as \textit{$LS_{(CNN|MB-DCT)}$}.

\section{Performance Evaluation}

\subsection{Experimental setup}\label{experiments}
We have experimented on the MNIST dataset that contains hand-written digit images and we followed the same experimental setup in \cite{cnn} as 50K and following 10K images are used in training and testing, respectively. %Firstly, we introduce the classifiers that we use in experiments. Then, we give details on implementation of all schemes. 

\paragraph{Computation of the features} To compute MB-DCT features, we learned a visual dictionary by K-means clustering based on hamming distance and using  training set consisting of 100 randomly selected  proxy images. We worked with K=512 clusters as in \cite{sinem}. The following procedures are as mentioned in Section IV.B.1. 

In order to compute $RS_{CNN}$ and $LS_{CNN}$ features we have re-implemented the corresponding architectures in \cite{cnn} and \cite{deeplearning} using the Keras library. For the $RS_{CNN}$ case, we have trained the network in \cite{cnn} using stochastic gradient descent with the parameters: learning rate 0.01, momentum 0.9, weight decay  0.0005, and we applied 15K epochs following \cite{cnn}. For the implementation of $LS_{CNN}$ we have trained the network in \cite{deeplearning} with Adam optimizer, using learning rate 0.00025 and 500 epochs. %For the both techniques we observed the convergence on the training set with the selected parameter values. 
Training took around 60 (due to high number of epochs) and 2 hours for the techniques of $RS_{CNN}$ and $LS_{CNN}$, respectively, with the GPU of GTX 1080 Ti. 

\paragraph{Choice of the classifier} We ran experiments with two different classifiers, namely, KNN and 1D-CNN. For KNN, we used the chi-square distance to compare histograms. We decided for the best value of 'k' by 5-fold cross-validation on the training set and then measured the  performance on the test set. 

We further wanted to examine the classification performance with a multilayer neural network. However, since the length of the features were quite high, i.e., 1012 for $RS_{CNN}$ and 596 for $LS_{CNN}$ (recall that these are also to be augmented with the 512 dimensional MB-DCT features in the fusion scheme), we decided not to follow this path to avoid excessive computational overhead. Instead we opted to train a 1D-CNN network, adopting Lenet-5 model, with the computed features of the training images. We used Adam optimizer with a learning rate of 0.00025 and 500 epochs in training which took around 2 hours for all the techniques.

\subsection{Performance results}
The performance results that are obtained with the aforementioned techniques in terms of test error are presented at Table \ref{tab2}. We also present three published performance results in the literature, namely, Smashed Filter \cite{smashedfilter3}, $RS_{CNN}$ \cite{cnn} and $LS_{CNN}$ \cite{deeplearning}. 

We observe that with our re-implementation of $RS_{CNN}$ and $LS_{CNN}$, we get performances quite close to the reported ones in \cite{cnn} and \cite{deeplearning}. For the degraded proxy image with random sampling, our binary descriptor ($RS_{MB-DCT}$) outperforms Smashed Filters \cite{smashedfilter3} significantly. $RS_{MB-DCT}$ also gives competitive results with respect to $RS_{CNN}$ \cite{cnn} for the lowest measurement rate (0.01). We outperform  $RS_{CNN}$ \cite{cnn} at the lowest measurement rate significantly when we use the fused feature (41.06\% vs 24.78\%).

The significant performance gain is achieved when degradation is created by the learned matrices. In that case, although $LS_{MB-DCT}$ was behind the reported $LS_{CNN}$ results in \cite{deeplearning}, our re-implementation of $LS_{CNN}$ was slightly better than theirs. More significantly, lowest classification error rates which can be accepted as the new state-of-the-art are obtained when we use joint features in $LS_{(CNN|MB-DCT)}$ implementation (3.87\% test error for 0.01 measurement rate).

%The results presented at Table \ref{tab2} are obtained by the KNN classifier. We also present the performance results obtained with 1D CNN at Table \ref{tab3}. 
The performance results presented in Table \ref{tab2} are obtained with the KNN classifier. We also present the performance results obtained with 1D-CNN at Table \ref{tab3}. As it can be seen in Table \ref{tab3}, although they were competitive for higher sampling rates, KNN always gives better result, more significantly at lowest measurement rate. However, execution time of KNN classifier was much higher than the 1D-CNN execution time on GPU.

\section{Conclusion}
In this work, we proposed a DCT-based discriminative feature scheme, computed directly from the proxy image which is usually a heavily degraded version of the original image. This feature vector (called MB-DCT) is binary, hence simple and low cost. We further introduced feature fusion by concatenating Bag of Words (BoW) representation of our binary features with one of the two state-of-the-art CNN-based feature vectors. Our experimental results show that proposed scheme gives the state-of-the-art performance for compressively sensed image classification even at the lowest measurement rate.

%In this work, we have proposed a new binary descriptor, called MB-DCT for image recognition tasks when used on inexactly reconstructed images, called proxy images. MB-DCT is not competitive with the state-of-the-art CNN features. However MB-DCT features prove their value in a feature-fusion scheme. Thus, when MB-DCT are fused (concatenated) with the CNN features, both having been extracted from proxy images reconstructed with the learned sensing matrix, they  outperform  all other feature combinations for practically all sensing rates.  Interestingly,  the lower the measurement rate, the more pronounced is the improvement due to fusion of learned and crafted features. 

%\newpage
\bibliography{ref}
\bibliographystyle{ieeetr}

\begin{comment}

\end{comment}

\end{document}